\documentstyle[aclap]{article}
\input{psfig}

\begin{document}

\title{\vspace{-0.5in}Learning Parse and Translation Decisions\\ From Examples With Rich Context}
\author{Ulf Hermjakob and Raymond J.\ Mooney\\
        Dept.\ of Computer Sciences\\
        University of Texas at Austin\\
        Austin, TX 78712, USA\\
        ulf@cs.utexas.edu \hspace{5mm} mooney@cs.utexas.edu}

\maketitle
\vspace{-0.5in}
\begin{abstract}
We present a knowledge and context-based system for parsing and translating
natural language and evaluate it on sentences from the Wall Street Journal.
Applying machine learning techniques, the system uses parse action examples
acquired under supervision to generate a deterministic shift-reduce parser
in the form of a decision structure. It relies heavily on context, as encoded in
features which describe the morphological, syntactic, semantic and
other aspects of a given parse state. 
\end{abstract}

\hyphenation{
in-ter-act-ive
cur-rent-ly
Mar-cin-kie-wicz
}
\bibliographystyle{fullname}

\section{Introduction}
The parsing of unrestricted text, with its enormous lexical and structural
ambiguity, still poses a great challenge in natural language processing.
The traditional approach of trying to master the complexity of parse grammars
with hand-coded rules turned out to be much more difficult than expected,
if not impossible.
Newer statistical approaches with often only very limited context sensitivity
seem to have hit a performance ceiling even when trained on very large corpora.

To cope with the complexity of unrestricted text, parse rules in any kind of formalism
will have to consider a complex context with many different 
morphological, syntactic or semantic features.
This can present a significant problem, because even linguistically trained natural
language developers have great difficulties writing and even more so extending explicit
parse grammars covering a wide range of natural language.
On the other hand it is much easier for humans to decide how {\it specific} sentences
should be analyzed. 

We therefore propose an approach to parsing based on learning from examples with
a very strong emphasis on context, integrating morphological, syntactic, 
semantic and other aspects relevant to making good parse decisions, thereby
also allowing the parsing to be deterministic.
Applying machine learning techniques, the system uses parse action examples
acquired under supervision to generate a deterministic shift-reduce type parser 
in the form of a decision structure.
The generated parser transforms input sentences into an integrated phrase-structure 
and case-frame tree, powerful enough to be fed into a transfer and a generation
module to complete the full process of machine translation.

Balanced by rich context and some background knowledge, 
our corpus based approach relieves the NL-developer from the hard if not impossible 
task of writing explicit grammar rules and keeps grammar coverage increases very 
manageable.
Compared with standard statistical methods, our system relies on deeper analysis
and more supervision, but radically fewer examples.

\section{Basic Parsing Paradigm}

As the basic mechanism for parsing text into a shallow semantic representation, 
we choose a shift-reduce type parser \cite{marc:mp80}.
It breaks parsing into an ordered sequence of small and manageable parse actions 
such as shift and reduce.
This ordered `left-to-right' parsing is much closer to how humans parse a sentence
than, for example, chart oriented parsers; it allows a very transparent control 
structure and makes the parsing process relatively intuitive for humans.
This is very important, because during the training phase,
the system is guided by a human supervisor for whom the flow of control needs 
to be as transparent and intuitive as possible.

The parsing does not have separate phases for part-of-speech selection and syntactic
and semantic processing, but rather integrates all of them into a single parsing phase. 
Since the system has all morphological, syntactic and semantic context information 
available at all times, the system can make well-based decisions very early, 
allowing a single path, i.e.\ deterministic parse, which eliminates wasting
computation on `dead end' alternatives. 

Before the parsing itself starts, the input string is segmented into a list of words 
incl.\ punctuation marks, which then are sent through a morphological analyzer that, 
using a lexicon\footnote{The lexicon provides part-of-speech information and links words
to concepts, as used in the KB (see next section). 
Additional information includes irregular forms and grammatical gender etc.\ (in the 
German lexicon).},
produces primitive frames for the segmented words.
A word gets a primitive frame for each possible part of speech.
(Morphological ambiguity is captured within a frame.)

\begin{figure}[htb]
\par
\centerline{\psfig{figure=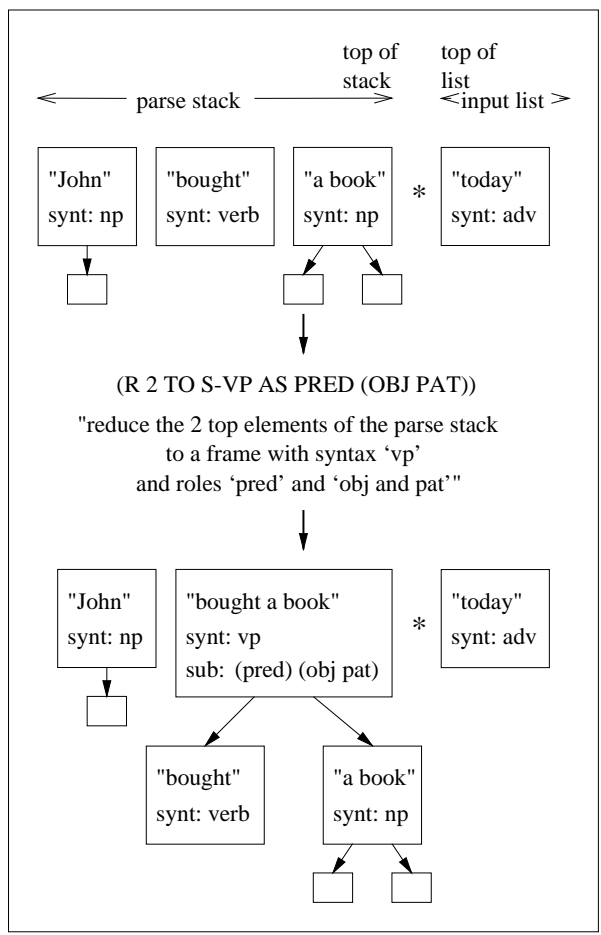,width=7.9cm}}
\par
\caption{Example of a parse action (simplified); boxes represent {\it frames}}
\label{fig_reduce_action}
\end{figure}

\begin{figure}[htb]
\begin{tabular}{|p{7.6cm}|} \hline
\begin{small}
\vspace{-1mm}
\begin{verbatim}
"John bought a new computer science book 
 today.":
   synt/sem: S-SNT/I-EV-BUY
   forms: (3rd_person sing past_tense)
   lex: "buy"
   subs:    
   (SUBJ AGENT) "John":
      synt/sem: S-NP/I-EN-JOHN
      (PRED)  "John"
         synt/sem: S-NOUN/I-EN-JOHN
   (PRED) "bought":
      synt/sem: S-TR-VERB/I-EV-BUY
   (OBJ THEME) "a new computer science book":
      synt/sem: S-NP/I-EN-BOOK
      (DET) "a"
      (MOD) "new"
      (PRED) "computer science book"
         (MOD) "computer science"
            (MOD) "computer"
            (PRED) "science"
         (PRED) "book"
   (TIME) "today":
      synt/sem: S-ADV/C-AT-TIME
      (PRED) "today"
         synt/sem: S-ADV/I-EADV-TODAY
   (DUMMY) ".":
      synt: D-PERIOD
\end{verbatim}
\end{small}\\ \hline
\end{tabular}
\caption{Example of a parse tree (simplified).}
\label{fig_parse_tree}
\end{figure}

The central data structure for the parser consists of a parse stack and an input list.
The parse stack and the input list contain trees of frames of words or phrases. 
Core slots of frames are surface and lexical form, syntactic and semantic category, 
subframes with syntactic and semantic roles, and
form restrictions such as number, person, and tense.
Optional slots include special information like the numerical value of number words.

Initially, the parse stack is empty and the input list contains the primitive frames
produced by the morphological analyzer. 
After initialization, the {\it deterministic parser} applies a sequence of 
{\it parse actions} to the parse structure. 
The most frequent parse actions are 
   {\it shift}, which shifts a frame from the input list onto the parse stack or backwards,
   and {\it reduce}, which combines one or several frames on the parse stack 
       into one new frame. The frames to be combined are typically, but not necessarily, 
       next to each other at the top of the stack. As shown in figure~\ref{fig_reduce_action}, 
       the action 

\begin{verbatim}
    (R 2 TO VP AS PRED (OBJ PAT))
\end{verbatim}

\noindent
for example reduces the two top frames of the stack into a new frame that is marked
as a verb phrase and contains the next-to-the-top frame as its predicate (or head)
and the top frame of the stack as its object and patient.
Other parse actions include {\it add-into}, which adds frames arbitrarily
deep into an existing frame tree, {\it mark}, which can mark any slot of any frame with 
any value, and operations to introduce empty categories (i.e.\ traces and `PRO', as
in ``She$_{i}$ wanted PRO$_{i}$ to win.''). Parse actions can have numerous arguments,
making the {\it parse action language} very powerful.

The parse action sequences needed for training the system are acquired interactively.
For each training sentence, the system and the supervisor parse the sentence step by step,
with the supervisor entering the next parse action,
e.g.\ {\tt (R 2 TO VP AS PRED (OBJ PAT))}, and the system executing it, 
repeating this sequence until the sentence is fully parsed. 
At least for the very first sentence, the supervisor actually has to type in the
entire parse action sequence. 
With a growing number of parse action examples available, the system, 
as described below in more detail, can be trained using those previous examples.
In such a partially trained system, the parse actions are then proposed by the system using 
a parse decision structure which ``classifies'' the current context. 
The proper classification is the specific action or sequence of actions that 
(the system believes) should be performed next. During further training,
the supervisor then enters parse action commands by either confirming 
what the system proposes or overruling it by providing the proper action.
As the corpus of parse examples grows and the system is trained on more and more data, 
the system becomes more refined, so that the supervisor has to overrule the system
with decreasing frequency.
The sequence of correct parse actions for a sentence is then recorded in a log file.

\section{Features}

To make good parse decisions, a wide range of features at various degrees of 
abstraction have to be considered.
To express such a wide range of features, we defined a {\it feature language}. 
Parse features can be thought of as functions that map from partially parsed sentences
to a value. Applied to the target parse state of figure~\ref{fig_reduce_action},
the feature {\it(SYNT OF OBJ OF -1 AT S-SYNT-ELEM)}, for example, designates the
general syntactic class of the object of the first frame of the parse
stack\footnote{{\it S-SYNT-ELEM} designates the top syntactic level; since -1 is
{\it negative}, the feature refers to the 1st frame of the {\it parse stack}.
Note that the top of stack is at the right end for the parse stack.},
in our example {\it np}\footnote{If a feature is not defined in
a specific parse state, the feature interpreter assigns the special value 
{\it unavailable}.}.
So, features do not a priori operate on words or phrases, but only do so if
their description references such words or phrases, as in our example through
the {\it path} `OBJ OF -1'.

Given a particular parse state and a feature, the system can interpret the feature 
and compute its value for the given parse state, often using additional background
knowledge such as\vspace{-1mm}
\begin{enumerate}
\item A {\it knowledge base} (KB), which currently consists of a directed acyclic 
    graph of 4356 mostly semantic and syntactic concepts
    connected by 4518 {\it is-a} links, e.g.\ 
    ``$book_{noun-concept}$ is-a\\ $tangible-object_{noun-concept}$''. Most concepts
    representing words are at a fairly shallow level of the KB,
    e.g.\ under `tangible object', `abstract', `process verb', or `adjective', 
    with more depth used only in concept areas more relevant for
    making parse and translation decisions, such as temporal, spatial and animate 
    concepts.\footnote{Supported by acquisition tools, word/concept pairs 
    are typically entered into the lexicon and the KB at the same time, 
    typically requiring less than a minute per word or group of closely related words.}\vspace{-1mm}
\item A {\it subcategorization table} that describes the syntactic and semantic role
    structures for verbs, with currently 242 entries.\vspace{-1mm}
\end{enumerate}

The following representative examples,
for easier understanding rendered in English and not in feature language syntax,
further illustrate the expressiveness of the feature language:
\begin{itemize}\vspace{-1mm}
\item the general syntactic class of $frame_{-3}$ 
   (the third element of the parse stack): e.g.\ verb, adj, np,\vspace{-1mm}
\item whether or not the adverbial alternative of $frame_{1}$ 
   (the top element of the input list) is an adjectival degree adverb,\vspace{-1mm}
\item the specific finite tense of $frame_{-1}$, e.g.\ present tense,\vspace{-1mm}
\item whether or not $frame_{-1}$ contains an object,\vspace{-1mm}
\item the semantic role of $frame_{-1}$ with respect to $frame_{-2}$:
   e.g.\ agent, time; this involves pattern matching with corresponding entries
    in the verb subcategorization table,\vspace{-1mm}
\item whether or not $frame_{-2}$ and $frame_{-1}$ satisfy subject-verb agreement.\vspace{-1mm}
\end{itemize}

Features can in principal refer to any one or several elements on the parse
stack or input list, and any of their subelements, at any depth. 
Since the currently 205 features are supposed to bear some linguistic relevance, 
none of them are unjustifiably remote from the current focus of a parse state.

\begin{figure*}[t]
\par
\centerline{\psfig{figure=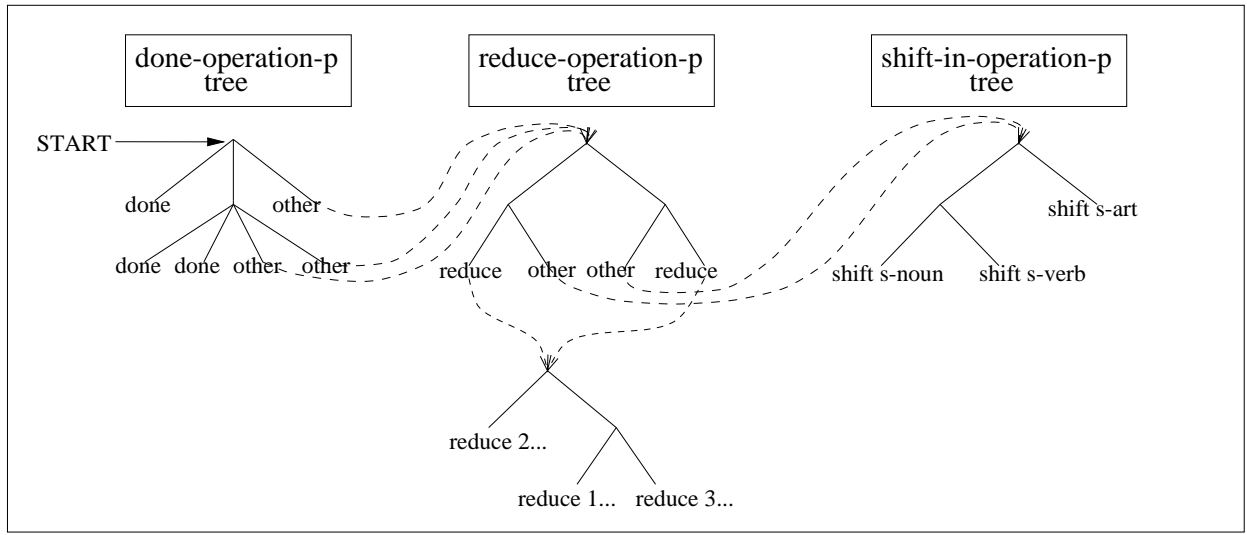,width=16.5cm}}
\par
\caption{Example of a hybrid decision structure}
\label{csg_dec_structure1}
\vspace{-1mm}
\end{figure*}

The feature collection is basically independent from the supervised parse action 
acquisition. Before learning a decision structure for the first time, the supervisor
has to provide an initial set of features that can be considered obviously relevant.
Particularly during the early development of our system, this set was increased whenever
parse examples had identical values for all current features but nevertheless demanded
different parse actions.
Given a specific conflict pair of partially parsed sentences, the supervisor would add
a new relevant feature that discriminates the two examples. We expect
our feature set to grow to eventually about 300 features when scaling up further within
the Wall Street Journal domain, and quite possibly to a higher number when expanding
into new domains. However, such feature set additions require fairly little supervisor 
effort.

Given (1) a log file with the correct parse action sequence of training sentences
as acquired under supervision and 
(2) a set of features, the system revisits the training sentences and computes values for
all features at each parse step. Together with the recorded parse actions these feature
vectors form {\it parse examples} that serve as input to the learning unit.
Whenever the feature set is modified, this step must be repeated,
but this is unproblematic, because this process is both fully automatic and fast.

\section{Learning Decision Structures}

Traditional statistical techniques also use features, but often have to sharply limit
their number (for some trigram approaches to three fairly simple features) to avoid
the loss of statistical significance.

In parsing, only a very small number of features are crucial over a wide range
of examples, while most features are critical in only a few examples, being used to
`fine-tune' the decision structure for special cases.
So in order to overcome the antagonism between the importance of having a large number of
features and the need to control the number of examples required for learning, particularly
when acquiring parse action sequence under supervision,
we choose a decision-tree based learning algorithm, which recursively selects the
most discriminating feature of the corresponding subset of training examples, eventually 
ignoring all locally irrelevant features, thereby tailoring the size of the final 
decision structure to the complexity of the training data.

While parse actions might be complex for the action interpreter, 
they are atomic with respect to the decision structure learner; 
e.g.\ ``(R 2 TO VP AS PRED (OBJ PAT))'' would be such an atomic {\it classification}.
A set of {\it parse examples}, as already described in the previous section,
is then fed into an ID3-based learning routine that generates 
a decision structure, which can then `classify' any given parse state
by proposing what parse action to perform next.

We extended the standard ID3 model \cite{quin:id6} to more general hybrid 
decision structures. In our tests, the best performing structure was a decision 
list \cite{rivest:ml87} of hierarchical decision trees, whose simplified basic structure 
is illustrated in figure~\ref{csg_dec_structure1}. 
Note that in the `reduce operation tree', the system first decides whether 
or not to perform a reduction before deciding on a specific reduction.
Using our knowledge of similarity of parse actions and the
exceptionality vs.\ generality of parse action groups, we can provide an overhead
structure that helps prevent data fragmentation.

\section{Transfer and Generation}

The output tree generated by the parser can be used for translation.
A transfer module recursively maps the source language parse tree to
an equivalent tree in the target language, reusing the methods developed 
for parsing with only minor adaptations. 
The main purpose of learning here is to resolve translation
ambiguities, which arise for example when translating the English ``{\it to know}''
to German ({\it wissen/kennen}) or Spanish ({\it saber/conocer}).

Besides word pair entries, the {\it bilingual dictionary} also contains
pairs of phrases and expressions in a format closely resembling traditional
(paper) dictionaries, e.g.\ ``to comment on SOMETHING\_1''/``sich zu ETWAS\_DAT\_1 \"{a}u{\ss}ern''.
Even if a complex translation pair does not bridge a structural mismatch, 
it can make a valuable contribution to disambiguation. 
Consider for example the term ``interest rate''. Both element nouns are
highly ambiguous with respect to German, but the English {\it compound} conclusively maps to 
the German compound ``Zinssatz''.
We believe that an extensive collection of complex translation pairs in the bilingual dictionary 
is critical for translation quality and we are confident that its acquisition can be at least 
partially automated by using techniques like those described in \cite{smadja:compling96}.
Complex translation entries are preprocessed using the same parser as for normal text.
During the transfer process, the resulting parse tree pairs are then accessed using pattern 
matching.

The generation module orders the components of phrases, adds appropriate
punctuation, and propagates morphologically relevant information in order 
to compute the proper form of surface words in the target language.

\section{Wall Street Journal Experiments}
\label{wsj-experiments}

We now present intermediate results on training and testing a prototype
implementation of the system with sentences from the Wall Street Journal,
a prominent corpus of `real' text, as collected on the ACL-CD.

In order to limit the size of the required lexicon, we work on a reduced corpus
of 105,356 sentences, a tenth of the full corpus,
that includes all those sentences that are fully covered by the 3000 most
frequently occurring words (ignoring numbers etc.) in the entire corpus.
The first 272 sentences used in this experiment vary in length from 4 to 45 words,
averaging at 17.1 words and 43.5 parse actions per sentence.
One of these sentence is ``{\it Canadian manufacturers' new orders fell to
\$20.80 billion (Cana-}

\begin{center}
\begin{tabular}{|l|c|c|c|c|c|} \hline \hline
   Tr.\ snt.                       & 16       & 32       & 64       & 128      & 256      \\ \hline \hline
   Prec.\                          & 85.1\%   & 86.6\%   & 87.7\%   & 90.4\%   & 92.7\%   \\
   Recall                          & 82.8\%   & 85.3\%   & 87.7\%   & 89.9\%   & 92.8\%   \\
   L.\ pr.\                        & 77.2\%   & 80.4\%   & 82.5\%   & 86.6\%   & 89.8\%   \\
   L.\ rec.                        & 75.0\%   & 77.7\%   & 81.6\%   & 85.3\%   & 89.6\%   \\
   Tagging\hspace{-0.5mm}          & 96.6\%   & 96.5\%   & 97.1\%   & 97.5\%   & 98.4\%   \\
   Cr/snt                          & 2.5      & 2.1      & 1.9      & 1.3      & 1.0      \\
   0 cr                            & 27.6\%   & 35.3\%   & 35.7\%   & 50.4\%   & 56.3\%   \\
   $\leq$ 1 cr                     & 44.1\%   & 50.7\%   & 54.8\%   & 68.4\%   & 73.5\%   \\
   $\leq$ 2 cr                     & 61.4\%   & 65.1\%   & 66.9\%   & 80.9\%   & 84.9\%   \\
   $\leq$ 3 cr                     & 72.4\%   & 77.2\%   & 79.4\%   & 87.1\%   & 93.0\%   \\
   $\leq$ 4 cr                     & 84.9\%   & 86.4\%   & 89.7\%   & 93.0\%   & 94.9\%   \\
   Ops                             & 79.1\%   & 82.9\%   & 86.8\%   & 89.1\%   & 91.7\%   \\
   OpSeq                           &  1.8\%   &  4.4\%   &  5.9\%   & 10.7\%   & 16.5\%   \\
   Str\&L                          &  5.5\%   &  8.8\%   & 10.3\%   & 18.8\%   & 26.8\%   \\
   Loops                           &   13     &   6      &    0     &    1     &    1     \\ \hline \hline
\end{tabular}
\end{center}
\noindent Table 1: Evaluation results with varying number of training sentences; 
  with all 205 features and hybrid decision structure;
  Train. = number of training sentences; pr\./prec.\ = precision; rec.\ = recall; l.\ = labeled;
  Tagging = tagging accuracy; Cr/snt = crossings per sentence; Ops = correct operations;
  OpSeq = Operation Sequence
\addtocounter{table}{1}

\vspace{5mm}
\centerline{\psfig{figure=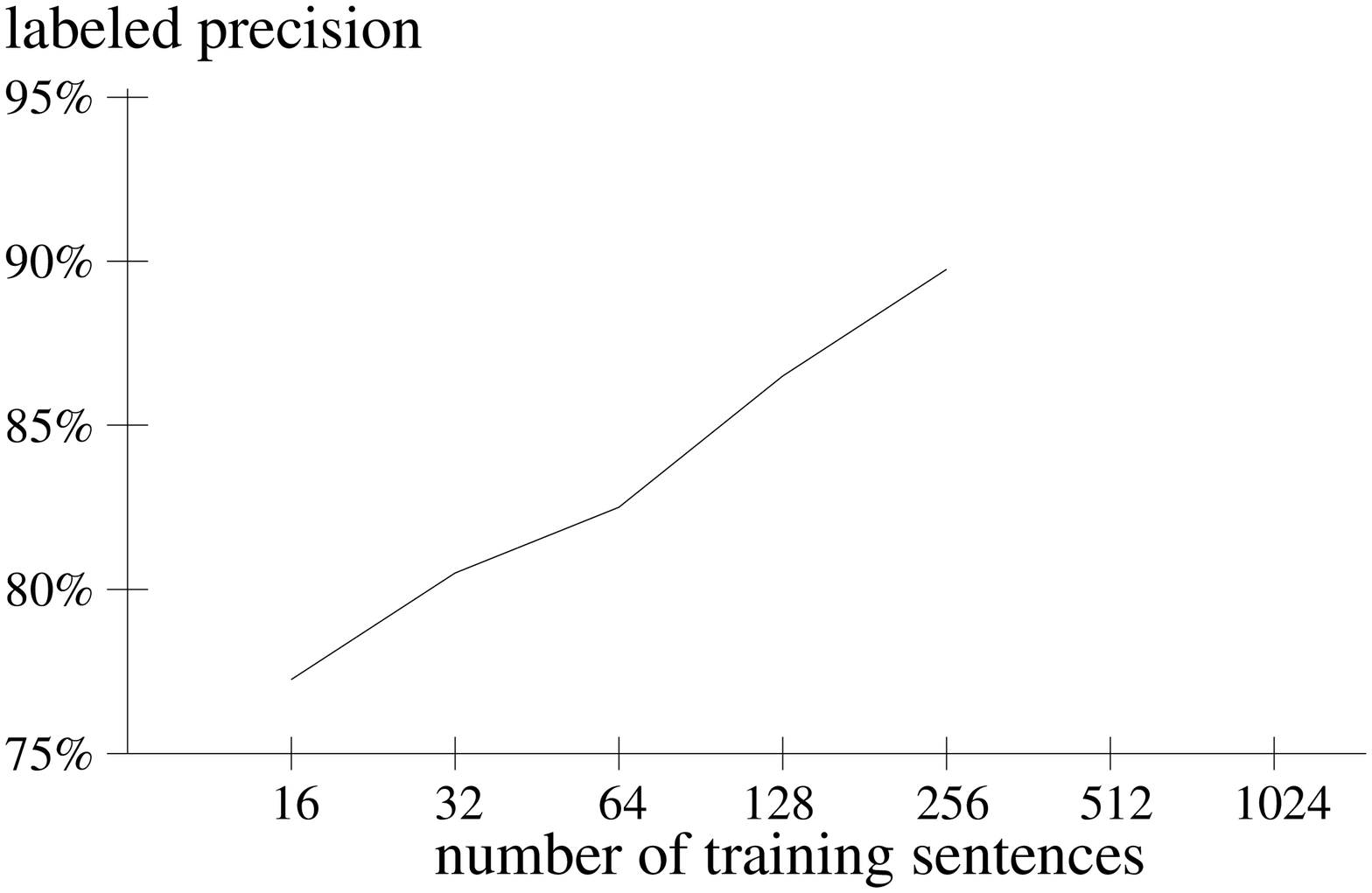,width=7.9cm}}
\noindent Figure 4: Learning curve for labeled precision in table 1.\\

\begin{table}[tb]
\begin{center}
\begin{tabular}{|l|c|c|c|c|c|} \hline \hline
   Features\hspace{-0.5mm}         & 6        & 25       &  50      & 100      & 205      \\ \hline \hline
   Prec.\                          & 88.0\%   & 88.7\%   & 90.8\%   & 91.7\%   & 92.7\%   \\
   Recall                          & 87.3\%   & 88.7\%   & 90.8\%   & 91.7\%   & 92.8\%   \\
   L.\ pr.\                        & 79.8\%   & 86.7\%   & 87.2\%   & 88.6\%   & 89.8\%   \\
   L.\ rec.                        & 81.5\%   & 84.1\%   & 86.9\%   & 88.1\%   & 89.6\%   \\
   Tagging\hspace{-0.5mm}          & 97.6\%   & 97.9\%   & 98.1\%   & 98.2\%   & 98.4\%   \\
   Cr/snt                          & 1.8      & 1.7      & 1.3      & 1.1      & 1.0      \\
   0 cr                            & 39.0\%   & 43.4\%   & 50.4\%   & 54.0\%   & 56.3\%   \\
   $\leq$ 1 cr                     & 57.4\%   & 59.6\%   & 70.6\%   & 72.1\%   & 73.5\%   \\
   $\leq$ 2 cr                     & 72.1\%   & 73.9\%   & 80.5\%   & 84.2\%   & 84.9\%   \\
   $\leq$ 3 cr                     & 82.7\%   & 84.9\%   & 88.6\%   & 92.3\%   & 93.0\%   \\
   $\leq$ 4 cr                     & 89.0\%   & 89.7\%   & 93.8\%   & 94.5\%   & 94.9\%   \\
   Ops                             & 80.6\%   & 81.9\%   & 88.9\%   & 90.7\%   & 91.7\%   \\
   OpSeq                           &  2.6\%   &  1.5\%   &  8.8\%   & 13.6\%   & 16.5\%   \\
   Str\&L                          &  8.8\%   &  9.2\%   & 15.1\%   & 23.5\%   & 26.8\%   \\
   Loops                           &    8     &   2      &    2     &    2     &    1     \\ \hline \hline
\end{tabular}
\caption{Evaluation results with varying number of features; with 256 training sentences}
\label{fig-var_feat}
\end{center}
\end{table}

\noindent {\it dian) in January, down 4\% from December's \$21.67 billion
billion on a seasonally adjusted basis, Statistics Canada, a federal agency, said.}''.

For our parsing test series, we use 17-fold cross-validation.
The corpus of 272 sentences that currently have parse action logs associated with them
is divided into 17 blocks of 16 sentences each. The 17 blocks are then consecutively
used for testing. For each of the 17 sub-tests, a varying number of
sentences from the {\it other} blocks is used for training the parse decision structure,
so that within a sub-test, none of the training sentences are ever used as a test sentence.
The results of the 17 sub-tests of each series are then averaged.\\

\vspace{-5mm}
\noindent{\bf Precision ({\it pr.}):} \hfill\\[-5mm]
\begin{flushright}
\begin{tabular}{c}
number of correct constituents in system parse \\ \hline
number of constituents in system parse \\
\end{tabular}\\
\end{flushright}
{\bf Recall ({\it rec.}):} \hfill\\[-5mm]
\begin{flushright}
\begin{tabular}{c}
number of correct constituents in system parse \\ \hline
number of constituents in logged parse \\
\end{tabular}\\
\end{flushright}

{\bf Crossing brackets} ({\it cr}): number of constituents which violate constituent boundaries 
with a constituent in the logged parse.\\
{\bf Labeled} ({\it l.}) precision/recall measures not only structural correctness, 
but also the correctness of the syntactic label.
{\bf Correct operations} ({\it Ops}) measures the number of correct operations during a parse that is
continuously corrected based on the logged sequence. The {\it correct operations} ratio is important 
for example acquisition, because it describes the percentage of parse actions that the supervisor can 
confirm by just hitting the return key.
A sentence has a correct {\bf operating sequence} ({\it OpSeq}), 
if the system fully predicts the logged parse action sequence,
and a correct {\bf structure and labeling} ({\it Str\&L}), if the structure 
and syntactic labeling of the final system parse of a sentence is 100\% correct, 
regardless of the operations leading to it.

\begin{table}[tb]
\begin{center}
\begin{tabular}{|l|c|c|c|c|c|} \hline \hline
   Type of deci-                   & plain    & hier.    & plain    & hybrid   \\ 
   sion structure                  & list     & list     & tree     & tree     \\ \hline
   Precision                       & 87.8\%   & 91.0\%   & 87.6\%   & 92.7\%   \\
   Recall                          & 89.9\%   & 88.2\%   & 89.7\%   & 92.8\%   \\
   Lab.\ precision                 & 28.6\%   & 87.4\%   & 38.5\%   & 89.8\%   \\
   Lab.\ recall                    & 86.1\%   & 84.7\%   & 85.6\%   & 89.6\%   \\
   Tagging acc.                    & 97.9\%   & 96.0\%   & 97.9\%   & 98.4\%   \\
   Crossings/snt                   & 1.2      & 1.3      & 1.3      & 1.0      \\
   0 crossings                     & 55.2\%   & 52.9\%   & 51.5\%   & 56.3\%   \\
   $\leq$ 1 crossings              & 72.8\%   & 71.0\%   & 65.8\%   & 73.5\%   \\
   $\leq$ 2 crossings              & 82.7\%   & 82.7\%   & 81.6\%   & 84.9\%   \\
   $\leq$ 3 crossings              & 89.0\%   & 89.0\%   & 90.1\%   & 93.0\%   \\
   $\leq$ 4 crossings              & 93.4\%   & 93.4\%   & 93.4\%   & 94.9\%   \\
   Ops                             & 86.5\%   & 90.3\%   & 90.2\%   & 91.7\%   \\
   OpSeq                           & 12.9\%   & 11.8\%   & 13.6\%   & 16.5\%   \\
   Str\&L                          & 22.4\%   & 22.8\%   & 21.7\%   & 26.8\%   \\
   Endless loops                   &   26     &  23      &   32     &    1    \\ \hline \hline
\end{tabular}
\caption{Evaluation results with varying types of decision structures;
        with 256 training sentences and 205 features}
\label{fig-var_dec_struct}
\end{center}
\end{table}

The current set of 205 features was sufficient to always discriminate examples
with different parse actions, resulting in a 100\% accuracy on sentences
already seen during training.
While that percentage is certainly less important than the accuracy figures for
unseen sentences, it nevertheless represents an important upper ceiling.

Many of the mistakes are due to encountering constructions that just have not been 
seen before at all, typically causing several erroneous parse decisions in a row.
This observation further supports our expectation, based on the results shown in table~1 
and figure~4, that with more training sentences,
the testing accuracy for unseen sentences will still rise significantly.

Table 2 shows the impact of reducing the feature set to a set of N core features. 
While the loss of a few specialized features will not cause a major degradation, the relatively 
high number of features used in our system finds a clear justification when evaluating compound
test characteristics, such as the number of structurally completely correct sentences.
When 25 or fewer features are used, all of them are syntactic. Therefore the 25 feature test
is a relatively good indicator for the contribution of the semantic knowledge base.

In another test, we deleted all 10 features relating to the
subcategorization table and found that the only metrics with degrading values were those measuring
semantic role assignment; in particular, none of the precision, recall and crossing bracket values
changed significantly. This suggests that, at least in the presence of other semantic features,
the subcategorization table does not play as critical a role in resolving structural ambiguity 
as might have been expected.

Table~\ref{fig-var_dec_struct} compares four different machine learning variants: plain decision lists,
hierarchical decision lists, plain decision trees and a hybrid structure, 
namely a decision list of hierarchical decision trees, as sketched in 
figure~\ref{csg_dec_structure1}. The results show that extensions to the basic
decision tree model can significantly improve learning results.

\begin{table}[htb]
\begin{center}
\begin{tabular}{|l|c|c|} \hline \hline
   System                          & Syntax  & Semantics \\ \hline 
   Human translation               & 1.18     & 1.41     \\
   {\sc Contex} on correct parse   & 2.20     & 2.19     \\
   {\sc Contex} (full translation) & 2.36     & 2.38     \\ \hline
   Logos                           & 2.57     & 3.24     \\
   {\sc Systran}                   & 2.68     & 3.35     \\
   Globalink                       & 3.30     & 3.83     \\ \hline \hline
\end{tabular}
\end{center}
\caption{Translation evaluation results (best possible = 1.00, worst possible = 6.00)}
\label{fig-trans_res}
\end{table}

Table~\ref{fig-trans_res} summarizes the evaluation results of translating 32 
randomly selected sentences from our Wall Street Journal corpus from English to German.
Besides our system, {\sc Contex}, we tested three commercial systems, Logos, {\sc Systran},
and Globalink.  In order to better assess the contribution of the parser, we also 
added a version that let our system
start with the correct parse, effectively just testing the transfer and generation module.
The resulting translations, in randomized order and without identification, were 
evaluated by ten bilingual graduate students, both native
German speakers living in the U.S. and native English speakers teaching college level German.
As a control, half of the evaluators were also given translations by a bilingual human.

Note that the translation results using our parser are fairly close to those 
starting with a correct parse.
This means that the errors made by the parser have had a relatively moderate impact
on translation quality. The transfer and generation modules were developed and trained
based on only 48 sentences, so we expect a significant translation quality improvement 
by further development of those modules.

Our system performed better than the commercial systems, but this has to
be interpreted with caution, since our system was trained and tested on sentences 
from the same lexically limited corpus (but of course without overlap), whereas 
the other systems were developed on and for texts from a larger variety of domains, 
making lexical choices more difficult in particular.

\begin{table}[htb]
\begin{center}
\begin{tabular}{|l|c|c|} \hline \hline
Metric               & Syntax & Semantics \\ \hline
Precision            &  -0.63 & -0.63  \\
Recall               &  -0.64 & -0.66  \\
Labeled precision    &  -0.75 & -0.78  \\
Labeled recall       &  -0.65 & -0.65  \\
Tagging accuracy     &  -0.66 & -0.56  \\
Number of crossing brackets\hspace*{-1mm} &  \hspace*{1mm}0.58 &  \hspace*{1mm}0.54  \\
Operations           &  -0.45 & -0.41  \\
Operation sequence   &  -0.39 & -0.36  \\ \hline  \hline
\end{tabular}
\end{center}
\caption{Correlation between various parse and translation metrics. Values near -1.0 or 1.0
indicate very strong correlation, whereas values near 0.0 indicate a weak or no correlation.
Most correlation values, incl.\ for labeled precision are negative, because a higher (better)
labeled precision correlates with a numerically lower (better) translation score on the
1.0 (best) to 6.0 (worst) translation evaluation scale.}
\label{fig-parse_trans_corr}
\end{table}

Table~\ref{fig-parse_trans_corr} shows the correlation between various parse and translation
metrics. Labeled precision has the strongest
correlation with both the syntactic and semantic translation evaluation grades.

\vspace{2.5mm}
\section{Related Work}

Our basic parsing and interactive training paradigm is based on \cite{simm:acl92}. 
We have extended their work by significantly increasing the expressiveness of the 
parse action and feature languages, in particular by moving far beyond the few simple
features that were limited to syntax only,
by adding more background knowledge and by introducing
a sophisticated machine learning component.

\cite{magerman:acl95} uses a decision tree model similar to ours, training his system
{\sc Spatter} with parse action sequences for 40,000 Wall Street Journal sentences 
derived from the Penn Treebank \cite{marcus:compling}.
Questioning the traditional n-grams, Magerman already advocates a heavier reliance on
contextual information. Going beyond Magerman's still relatively rigid set of 36 features, 
we propose a yet richer, basically unlimited feature language set. Our parse action
sequences are too complex to be derived from a treebank like Penn's. Not only do our
parse trees contain semantic annotations, roles and more syntactic detail, we also rely
on the more informative parse action sequence. While this necessitates the involvement 
of a parsing supervisor for training, we are able to perform deterministic parsing
and get already very good test results for only 256 training sentences.

\cite{collins:acl96} focuses on bigram lexical dependencies ({\sc Bld}).
Trained on the same 40,000 sentences as {\sc Spatter},
it relies on a much more limited type of context than our system
and needs little background knowledge.

\begin{table}[htb]
\begin{center}
\begin{tabular}{|l|c|c|c|} \hline \hline
   Model                           & {\sc Spatter}  & {\sc Bld} & {\sc Contex}   \\ \hline \hline
   Labeled precision               & 84.9\%   & 86.3\%    & 89.8\%   \\
   Labeled recall                  & 84.6\%   & 85.8\%    & 89.6\%   \\
   Crossings/sentence              & 1.26     & 1.14      & 1.02     \\
   Sent.\ with 0 cr.               & 56.6\%   & 59.9\%    & 56.3\%   \\
   Sent.\ with $\leq$ 2 cr.\hspace*{-0.5mm} & 81.4\%   & 83.6\%   & 84.9\%   \\ \hline \hline
\end{tabular}\\[2mm]
\end{center}
\caption{Comparing our system {\sc Contex} with Magerman's {\sc Spatter} and Collins' {\sc Bld}; 
results for {\sc Spatter} and {\sc Bld} are for sentences of up to 40 words.}
\label{fig-rel_work}
\end{table}

Table~\ref{fig-rel_work} compares our results with {\sc Spatter} and {\sc Bld}.
The results have to be interpreted cautiously since they
are not based on the exact same sentences and detail of bracketing.
Due to lexical restrictions, our average sentence length (17.1) is below the one
used in {\sc Spatter} and {\sc Bld} (22.3),
but some of our test sentences have more than 40 words;
and while the Penn Treebank leaves many phrases such as ``the
New York Stock Exchange'' without internal structure,
our system performs a complete bracketing, thereby increasing the risk of crossing
brackets.

\section{Conclusion}

We try to bridge the gap between the typically hard-to-scale hand-crafted 
approach and the typically large-scale but context-poor statistical approach for 
unrestricted text parsing.

\noindent Using \vspace{-2mm}
\begin{itemize}
  \item a rich and unified context with 205 features, \vspace{-2.5mm}
  \item a complex parse action language that allows integrated
        part of speech tagging and syntactic and semantic processing, \vspace{-2.5mm}
  \item a sophisticated decision structure that generalizes traditional decision trees and
 lists, \vspace{-2.5mm}
  \item a balanced use of machine learning and {\it micromodular} background knowledge, 
   i.e.\ very small pieces of highly independent information\vspace{-2.5mm}
  \item a modest number of interactively acquired examples from the Wall Street Journal, \vspace{-2.5mm}
\end{itemize}
\noindent our system {\sc Contex} \vspace{-2.5mm}
\begin{itemize}
  \item computes parse trees and translations fast, because it uses a deterministic 
        single-pass parser, \vspace{-6.5mm}
  \item shows good robustness when encountering novel constructions, \vspace{-2.5mm}
  \item produces good parsing results comparable to those of the leading statistical methods, and \vspace{-2.5mm}
  \item delivers competitive results for machine translations.
\end{itemize}

While many limited-context statistical approaches have already reached a 
performance ceiling, we still expect to significantly improve our results when 
increasing our training base beyond the currently 256 sentences, because
the learning curve hasn't flattened out yet and adding substantially more 
examples is still very feasible.
Even then the training size will compare favorably with the
huge number of training sentences necessary for many statistical systems.


\vspace{-4.5mm}
\begin{thebibliography}{}
\renewcommand{\baselinestretch}{0.97}
\bibitem[\protect\citename{Black \bgroup et al.\egroup }1992]{black:stat92}
E. Black, J. Lafferty, and S. Roukos.
\newblock 1992.
\newblock Development and evaluation of a broad-coverage probabilistic grammar
of English-language computer manuals.
\newblock In {\em 30th Proceedings of the ACL}, pages 185--192.\vspace{-2.2mm}

\bibitem[\protect\citename{Collins}1996]{collins:acl96}
M.~J. Collins.
\newblock 1996.
\newblock A New Statistical Parser Based on Bigram Lexical Dependencies.
\newblock In {\em 34th Proceedings of the ACL}, pages 184--191.\vspace{-2.2mm}

\bibitem[\protect\citename{Hermjakob}1997]{hermjakob:thesis97}
U. Hermjakob.
\newblock 1997.
\newblock {\em Learning Parse and Translation Decisions From Examples With Rich Context}.
\newblock Ph.D.\ thesis, Univ.\ of Texas at Austin, Dept.\ of Computer Sciences TR 97-12.
\newblock \mbox{file:/\hspace*{-0.3mm}/ftp.cs.utexas.edu/\hspace*{-0.2mm}pub\hspace*{-0.3mm}/\hspace*{-0.3mm}mooney\hspace*{-0.2mm}/\hspace*{-0.2mm}papers/herm} jakob-dissertation-97.ps.Z\vspace{-2.2mm}

\bibitem[\protect\citename{Magerman}1995]{magerman:acl95}
D.~M. Magerman.
\newblock 1995.
\newblock Statistical Decision-Tree Models for Parsing
\newblock In {\em 33rd Proceedings of the ACL}, pages 276--283.\vspace{-2.2mm}

\bibitem[\protect\citename{Marcus}1980]{marc:mp80}
M.~P. Marcus.
\newblock 1980.
\newblock {\em A Theory of Syntactic Recognition for Natural Language}.
\newblock MIT Press.\vspace{-2.2mm}

\bibitem[\protect\citename{Marcus \bgroup et al.\egroup }1993]{marcus:compling}
M.~P. Marcus, B. Santorini, and M.~A. Marcinkiewicz.
\newblock 1993.
\newblock Building a Large Annotated Corpus of {English}: The {Penn} Treebank.
\newblock In {\em Computational Linguistics 19 (2)}, pp.\ 184--191.\vspace{-2.2mm}

\bibitem[\protect\citename{Nirenburg \bgroup et al.\egroup }1992]{cmu:mt92}
S. Nirenburg, J. Carbonell, M. Tomita, and K. Goodman.
\newblock 1992.
\newblock {\em Machine Translation: A Knowledge-Based Approach}.
\newblock San Mateo, CA: Morgan Kaufmann.\vspace{-2.2mm}

\bibitem[\protect\citename{Quinlan}1986]{quin:id6}
J.R.\ Quinlan.\
\newblock 1986.\
\newblock Induction of decision trees.\
\newblock In {\em Machine Learning 1 (1)}, pp.\ 81--106.\vspace{-2.2mm}

\bibitem[\protect\citename{Rivest}1987]{rivest:ml87}
R. L. Rivest.
\newblock 1987.
\newblock Learning Decision Lists.
\newblock In {\em Machine Learning 2}, pages 229--246.\vspace{-2.2mm}

\bibitem[\protect\citename{Simmons and Yu}1992]{simm:acl92}
R.~F. Simmons and Yeong-Ho Yu.
\newblock 1992.
\newblock The Acquisition and Use of Context-Dependent Grammars for English.
\newblock In {\em Computational Linguistics 18 (4)}, pp.\ 391--418.\vspace{-2.2mm}

\bibitem[\protect\citename{Smadja et al.}1996]{smadja:compling96}
F. Smadja, K. R. McKeown and V. Hatzivassiloglou.
\newblock 1996.
\newblock Translating Collocations for Bilingual Lexicons: A Statistical Approach.
\newblock In {\em Computational Linguistics 22 (1)}, pages 1--38.\vspace{-2.2mm}

\bibitem[\protect\citename{Globalink}]{com_mt:globalink96}
\newblock http://www.globalink.com/
\newblock {\em Oct.\ 1996}.\vspace{-2.2mm}

\bibitem[\protect\citename{Logos}]{com_mt:logos96}
\newblock http://www.logos-ca.com/
\newblock {\em Oct.\ 1996}.\vspace{-2.2mm}

\bibitem[\protect\citename{{\sc Systran}}]{com_mt:systran96}
\newblock http://systranmt.com/
\newblock {\em Oct.\ 1996}.

\end{thebibliography}
\end{document}